\documentclass{article}

\usepackage{microtype}
\usepackage{graphicx}
\usepackage{booktabs} 

\usepackage{hyperref}

\usepackage{xspace}
\usepackage{amsfonts}       
\usepackage{nicefrac}       
\usepackage{microtype}      
\usepackage{color}
\usepackage{amssymb,amsmath,bm,amsthm}
\usepackage{subcaption}
\usepackage[usenames,dvipsnames]{xcolor}  

\usepackage[]{todonotes}  
\usepackage{arydshln}
\usepackage{sidecap}
\usepackage{wrapfig}
\usepackage{enumitem}

\usepackage[accepted]{sysml2019}

\usepackage{listings}             

\renewcommand{\paragraph}[1]{\noindent\textbf{#1}\xspace\xspace}

\newcommand{\surrealsys}{\textsc{Surreal-System}\xspace}
\newcommand{\surreal}{\textsc{Surreal}\xspace}
\newcommand{\cloudwise}{\textsc{Cloudwise}\xspace}
\newcommand{\symphony}{\textsc{Symphony}\xspace}
\newcommand{\caraml}{\textsc{Caraml}\xspace}
\newcommand{\sddpg}{\textsc{\surreal-DDPG}\xspace}
\newcommand{\sppo}{\textsc{\surreal-PPO}\xspace}
\newcommand{\ses}{\textsc{\surreal-ES}\xspace}

\def\EE{\mathbb{E}}

\newcommand{\lb}{\left [}
\newcommand{\rb}{\right ]}

\definecolor{mygreen}{rgb}{0,0.6,0}
\definecolor{mygray}{rgb}{0.5,0.5,0.5}
\definecolor{mymauve}{rgb}{0.58,0,0.82}

\sysmltitlerunning{\surrealsys: Fully-Integrated Stack for Distributed Deep Reinforcement Learning}

\begin{document}

\twocolumn[
\sysmltitle{\surrealsys: Fully-Integrated Stack for \\Distributed Deep Reinforcement Learning}



\sysmlsetsymbol{equal}{*}

\begin{sysmlauthorlist}
\sysmlauthor{Linxi Fan}{equal}\qquad
\sysmlauthor{Yuke Zhu}{equal}\qquad
\sysmlauthor{Jiren Zhu}{}\qquad
\sysmlauthor{Zihua Liu}{}\qquad
\sysmlauthor{Orien Zeng}{}\\
\sysmlauthor{Anchit Gupta}{}\qquad
\sysmlauthor{Joan Creus-Costa}{}\qquad
\sysmlauthor{Silvio Savarese}{}\qquad
\sysmlauthor{Li Fei-Fei}{}\\
\url{https://surreal.stanford.edu/}
\end{sysmlauthorlist}


\sysmlcorrespondingauthor{Linxi Fan}{jimfan@cs.stanford.edu}
\sysmlcorrespondingauthor{Yuke Zhu}{yukez@cs.stanford.edu}

\sysmlkeywords{Distributed Machine Learning, Reinforcement Learning}

\vskip 0.3in

\begin{abstract}
We present an overview of \surrealsys, a reproducible, flexible, and scalable framework for distributed reinforcement learning (RL). The framework consists of a stack of four layers: Provisioner, Orchestrator, Protocol, and Algorithms. The Provisioner abstracts away the machine hardware and node pools across different cloud providers. The Orchestrator provides a unified interface for scheduling and deploying distributed algorithms by high-level description, which is capable of deploying to a wide range of hardware from a personal laptop to full-fledged cloud clusters. The Protocol provides network communication primitives optimized for RL. Finally, the \surreal algorithms, such as Proximal Policy Optimization (PPO) and Evolution Strategies (ES), can easily scale to 1000s of CPU cores and 100s of GPUs. The learning performances of our distributed algorithms establish new state-of-the-art on OpenAI Gym and Robotics Suites tasks.
\end{abstract}
]



\printAffiliationsAndNotice{\sysmlEqualContribution} 

\section{Introduction}


Distributed systems development is becoming increasingly important in the field of deep learning. Prior work has demonstrated the value of distributing computation to train deep networks with millions of parameters on large, diverse datasets~\cite{dean2012large, goyal2017accurate, Moritz2017}. Distributed learning systems have recently witnessed a great deal of success across a wide variety of games, tasks, and domains~\cite{dean2012large, silver2016mastering, goyal2017accurate, liang2017ray, horgan2018distributed, espeholt2018impala}. In particular, distributed Reinforcement Learning (RL) has demonstrated impressive state-of-the-art results across several sequential decision making problems such as video games and continuous control tasks.

 \begin{figure}[h!]
 \begin{center}
 \includegraphics[width=1.\linewidth]{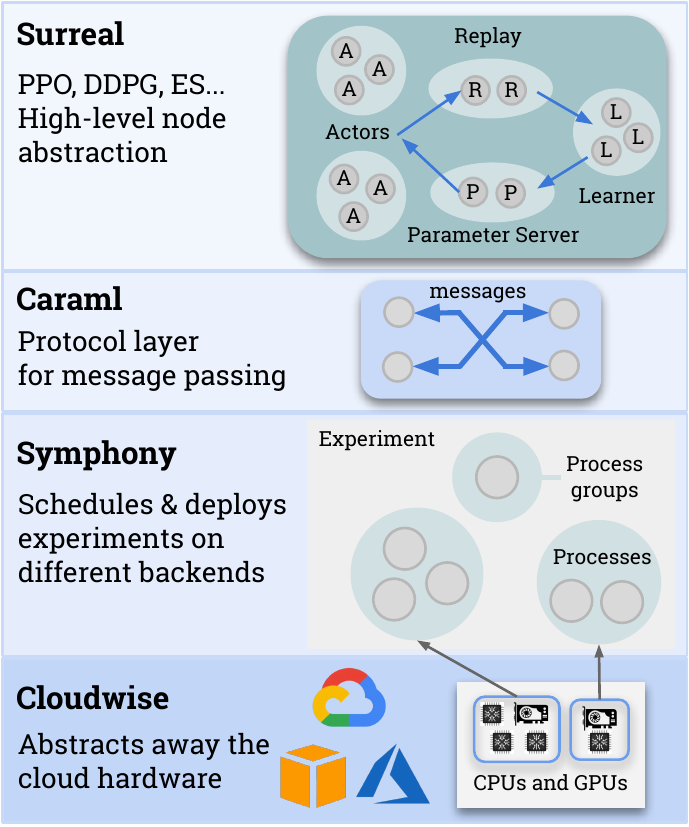}
  \vspace{-6mm}
  \caption{Overview of the four-layer stack in \surrealsys: \textsc{Cloudwise} for cloud provisioning, \textsc{Symphony} for container orchestration, \textsc{Caraml} as a communication layer, and \surreal for distributed reinforcement learning algorithms.}
 \label{fig:pull-fig}
 \end{center}
   \vspace{-7mm}
 \end{figure}

Three important design paradigms are desired for distributed reinforcement learning systems: reproducibility, flexibility, and scalability. Reproducing and validating prior deep reinforcement learning results is rarely straightforward, as it can be affected by numerous factors in the tasks and the underlying hardware. It makes comparing and evaluating different algorithms difficult~\cite{henderson2017deep}. In order to ensure progress in the field, a distributed reinforcement learning system must produce results that are easily \textit{reproducible} by others. Furthermore, to support a wide spectrum of algorithms and enable the creation of new algorithms, the distributed system must also be \textit{flexible}. Implementing a new RL algorithm should not require re-engineering intermediate system components or adding new ones. Instead, existing system components need to be able to adapt to the needs of upstream algorithms and use cases. Finally, the system should exhibit significant capability to \textit{scale} to leverage large quantities of physical computing resources efficiently. Deep RL methods often suffer from high sample complexity due to the burden of exploring large state spaces and can benefit from diverse sets of experience~\cite{horgan2018distributed}. Leveraging large-scale distributed computation can help these methods collect diverse experience quickly and help improve convergence rates.

Although existing distributed reinforcement learning algorithms such as Ape-X~\cite{horgan2018distributed} and IMPALA~\cite{espeholt2018impala} demonstrate an impressive capacity to scale up to many machines and achieve state-of-the-art results across a wide spectrum of Atari benchmarks and continuous control domains, these results are not easily reproducible by other researchers due to the difficulty of reimplementing the software infrastructure and gathering the hardware resources necessary to recreate their experiments. Many existing open-source reinforcement learning frameworks aim at providing high-quality implementations of these standard distributed RL algorithms, but they do not provide fully-integrated support for the corresponding hardware resources~\cite{baselines, hafner2017tensorflow, liang2017ray}.

Other existing systems in deep learning have also addressed some of these design paradigms. TVM~\cite{chen2018tvm} is an open-source end-to-end optimizing compiler for maintaining deep learning model performance across different hardware backends, while VTA~\cite{moreau2018vta} builds on top of the TVM compiler to provide an entire deep learning stack that allows for control over both high-level software and low-level hardware optimizations. While neither of these frameworks support scalability, the end-to-end control of both hardware and software allows for both reproducibility of model performance and a flexible approach to designing models that can leverage different types of hardware capabilities.

In order to establish a distributed RL framework that produces consistent and reproducible results that can be verified and built upon by researchers, the framework must provide an end-to-end solution, from hardware provisioning and deployment to algorithms that operate at the highest level of abstraction. Enabling control over hardware resources can benefit the flexibility and scalability of such a system as well by supporting algorithms that might require several computing nodes running on heterogenous hardware and software specifications. Therefore, we propose a general open-source sequential decision making framework that provides the entire stack --- from hardware deployment to algorithms.

We present \surrealsys, a fully-integrated stack for distributed deep reinforcement learning. \surrealsys consists of four primary layers. The first layer, the Provisioner (\cloudwise library), offers a common hardware abstraction for instance types and node pools across different cloud vendors. It prepares the foundation for the next layer, the Orchestrator (\symphony library), that builds upon a community-standard cloud API (Kubernetes). \symphony allows users to specify an experiment's launch logic, hardware resources, and network connectivity patterns in high-level description. It provides a unified frontend for orchestration and deployment to different backends ranging from local laptop to cloud clusters. Once an experiment has been configured, the Protocol layer (\caraml library) handles all the communications between algorithmic components. Simulated experience data and neural network parameters can be transferred very efficiently, thanks to the RL-specific optimizations we make. Finally, we provide competitive implementations of policy gradient algorithms and Evolution Strategies (\surreal libary) as well as demonstrating their performance and capacity to scale in large-scale experiments.




In Sec.~\ref{sec:benchmark} we show how our stack scales efficiently to heterogeneous clusters having 1000s of CPUs and 100s of GPUs when using both Proximal Policy Optimization(PPO) and Evolution Strategies (ES). We achieve near linear scaling in environment frames per second with number of agents demonstrating the effectiveness of our communication stack. We also show how optimizations like a load balanced replay, batched actors and optimized communication help us further improve system performance.

We also show the learning performance of our algorithms in Sec.~\ref{sec:exp} for a range of OpenAI Gym tasks as well as more challenging Robotics Suite tasks introduced in our prior work~\cite{fan2018surreal}. By using 1024 agents, our PPO implementation is able to partially solve even the most challenging tasks which were unsolved when using fewer agents hence showing the advantage of massively distributed RL. Our ES implementation also outperforms the reference implementation~\cite{liang2017ray} on a range of OpenAI Gym tasks. The contributions of our work are as follows:

\begin{enumerate}
  \item We propose \surrealsys, a reproducible, flexible, and scalable framework for distributed reinforcement learning.
  \item We introduce a streamlined four-layer stack that provides an end-to-end solution from hardware to algorithms. It can both massively scale on full-fledged cloud clusters and quickly iterate research ideas on a local machine.
  \item We describe the details and perform evaluations of our algorithmic implementations on a variety of control tasks. They show highly competitive with the state-of-the-art results.
\end{enumerate}

\section{\surrealsys: Distributed Reinforcement Learning Stack}


\begin{table*}[t]
    \centering
\caption{\textsc{Symphony} backend feature comparisons.}
\begin{tabular}{llccc}
\hline
Backends       & Scalability                & Autoscaling & Containerized & Debugging \\ \hline
\textit{Kubernetes}     & Multi-node on cloud                 & Yes                  & Yes                    & Slow               \\
\textit{Tmux}           & Local only                          & No                   & No                     & Fast               \\
\textit{Docker-compose} & Multi-node only with \texttt{Swarm} & No                   & Yes                    & Fast               \\
\textit{Minikube}       & Local only, simulates cloud         & No                   & Yes                    & Slow               \\
\hline
\end{tabular}
\label{tab:symphony-backends}
\end{table*}

Unlike data parallelism in supervised learning, distributed reinforcement learning (RL) algorithms require complex communication patterns between heterogeneous components, increasing the burden of engineering in reinforcement learning research. Our goal is to open source a computing infrastructure that makes the runtime setup effortless to upper-level algorithm designers with \emph{flexibility}, \emph{reproducibility} and \emph{scalability} as our guiding principles.  

We design a four-layer distributed learning stack as shown in Fig.~\ref{fig:pull-fig}, which decouples the end-user algorithms from the underlying computing runtime. Users of \surrealsys should be able to replicate our cluster setup, reproduce our experimental results, and rapidly iterate new research ideas on top of our algorithmic libraries. Each component  can be used independent of each other, or even outside the \surreal context to facilitate other cloud-based distributed tasks. Here we provide an overview of our four-layer stack for distributed RL systems, from the cloud computing hardware to the RL algorithm implementations.

\subsection{Provisioner: \textsc{Cloudwise}}

 \textsc{Cloudwise} aims to achieve a fully reproducible \surreal cluster setup running on the users' cloud account. \textsc{Cloudwise} abstracts away the cloud instances and node-pool configurations between different cloud providers. For example, Google Cloud machine type identifiers look like \texttt{n1-standard-4} and \texttt{n1-highmem-64}, while Amazon AWS uses a distinct naming scheme such as \texttt{t2.small} and \texttt{m5d.24xlarge}. They also have different mechanisms to deploy native clusters.

To standardize across different conventions, \textsc{Cloudwise} makes the following two design choices:

\begin{enumerate}
  \item Lifts a cloud account to be Kubernetes-ready. Kubernetes~\cite{burns2016borg} is a well-established, open-source cloud API standard compatible with all major cloud providers.
  \item Uses descriptive Python parameters (e.g.~\texttt{cpu=48, gpu\_type="v100", gpu=4}) that translate to the corresponding terminology on different cloud services.
\end{enumerate}

\textsc{Cloudwise} automates the tedious and convoluted procedure of setting up a customized \surreal Kubernetes cluster from scratch. The library only needs to be run once before any experiment. After the setup,
the upper layers will not be able to tell apart the differences between cloud providers, such as GCE or AWS. \textsc{Cloudwise} is also capable of adding, removing, and editing the properties of node pools after the cluster has been created.

\subsection{Orchestrator: \textsc{Symphony}}

Once the user sets up the cloud account with \textsc{Cloudwise}, the \textsc{Symphony} library takes over and helps orchestrate \texttt{Experiment}s on top of Kubernetes.

Each \texttt{Experiment} is a logical set of \texttt{Process}es and \texttt{ProcessGroup}s that communicate with each other through intra-cluster networking. A \texttt{ProcessGroup} contains a number of \texttt{Process}es that are guaranteed to be scheduled on the same physical node. Users can easily specify per-process resources (e.g., CPU-only node for actors and GPU nodes for learner) as well as their network connectivity, while \textsc{Symphony} takes care of scheduling and book-keeping. It uses containerization (Docker) to ensure that the runtime environment and dependencies are reproducible.

Furthermore, \textsc{Symphony} supports auto-scaling out of the box. The auto-scaling mechanism spins up or tears down nodes as new experiments start or old experiments terminate. It is highly economical for small research groups because no cloud instances are left running without a workload. Team members can collaborate on the same cluster and check each others' experiment status.

\subsubsection{Multiple Orchestration Backends}

A prominent feature of \textsc{Symphony} is its flexibility to deploy the same experiment logic on different computing environments, ranging from a personal laptop to a full-fledged cloud cluster. In this regard, \textsc{Symphony} provides the users with a unified interface for process orchestration and networking. Once users specify an abstract \texttt{Experiment} and connectivity configuration, they can choose from a variety of backends that satisfy different stage of development. We currently support the following four modes (also summarized in Table~\ref{tab:symphony-backends}):

\paragraph{Kubernetes (``Kube") backend.} This is the primary use case of \textsc{Symphony}. It deploys to the Kubernetes engine on the users' cloud provider, and its scalability is limited only by the resource quota and budget. All processes (``pods" in Kubernetes terminology) are containerized, thus avoiding dependency issues. 
Kubernetes is suitable for deploying large-scale experiments after validating on a smaller scale. To enable fast iteration and development, we provide Tmux, Docker-compose and Minikube as well.

\paragraph{Tmux backend.} \texttt{tmux} is a terminal multiplexer that allows users to access multiple terminal sessions in separate ``panes" inside a single window. We build the backend upon \texttt{tmux}'s panes to deploy a distributed experiment on an interactive machine (personal laptop/desktop or ssh-reachable machine). The merit is that we can now quickly iterate on the codebase. Any code updates will be reflected immediately in the local \texttt{tmux} session; error messages, if any, will emerge almost instantly. The demerit is that there is no containerization, and users will be responsible for installing the dependencies manually. In addition, \texttt{tmux} does not support multi-node distributed training.

\paragraph{Docker-compose backend.} This backend is a reconciliation of Tmux and Kube modes. It deploys locally on any interactive machine that has \texttt{docker} installed. \texttt{Docker-compose} eliminates the need to install complex dependencies locally and does not incur the Kube pod creation overhead. In future upgrades, we will also support \texttt{docker-swarm}, which will enable multi-node communication for this backend.

\paragraph{Minikube backend.} Conceptually the same as the Kube backend, \texttt{minikube} deploys on a local machine and simulates the full-blown cloud environment. Users can test their implementation thoroughly before running on the cloud account to avoid any unnecessary computing costs. Minikube backend has exactly the same API and command line interface as Kube backend, thus the  migration to the real cluster will be relatively effortless.

\subsubsection{Load-balancing and Sharding}
\label{sec:symphony-scalability}
Kubernetes is designed to provide horizontal scalability and handle large workloads. When using \symphony and \caraml on Kubernetes, components of a distributed learning algorithm can be sharded and requests to them are load balanced. For example, when a single replay server is not able to handle data generated by a large number of actors, one can create multiple shards. \symphony manages service declaration on Kubernetes to ensure that they process workload evenly. The effect of sharding the replay server is further discussed in Sec.~\ref{sec:sharding-replays}.

\subsection{Protocol: \textsc{Caraml}}

The \textsc{Caraml} library (\textit{CAR}efree \textit{A}ccelerated \textit{M}essaging \textit{L}ibrary) is our communication protocol based on ZeroMQ (a high performance messaging library) and Apache Arrow (in-memory data format for fast serialization). \textsc{Caraml} implements highly scalable distributed directives, such as \texttt{Push-Pull} for actors sending experience to replay and \texttt{Publish-Subscribe} for broadcasting parameters to actors. \textsc{Caraml} offers a more transparent alternative to frameworks like distributed TensorFlow~\cite{abadi2016tensorflow}, and applies to use cases beyond machine learning as well.

In distributed learning scenarios, high-dimensional observation vectors and large network parameters frequently make communication and serialization the bottleneck of the system. \caraml provides specific optimizations to avoid these problems. For example, we use \caraml Data Fetcher to fetch data from separate processes (actors) and transfer to the main process using a shared memory (reply buffer). This offloads serialization and networking from the main process. We quantitatively examine the speed-up of RL algorithms from \caraml's optimizations in Sec.~\ref{sec:caraml-speed-ablation}.


\begin{figure*}
        \centering
        \hfill
\begin{subfigure}[b]{0.5\linewidth}

                \includegraphics[width=\textwidth]{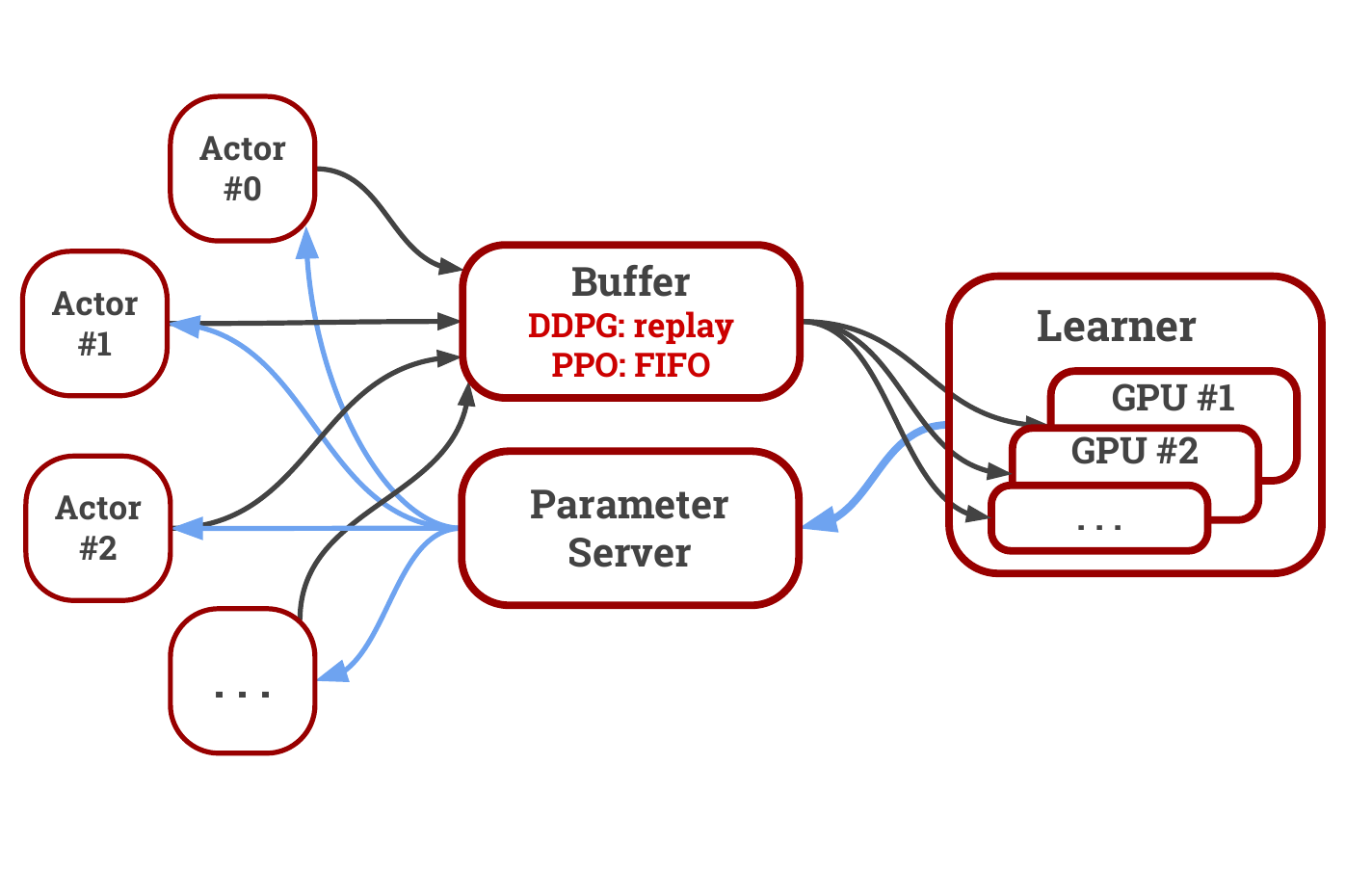}
                \caption{Distributed System Diagram for Reinforcement Learning}
                \label{fig:ppoddpg-diag}
        \end{subfigure}
        \hfill
\begin{subfigure}[b]{0.45\linewidth}
            \centering
                \includegraphics[width=\textwidth]{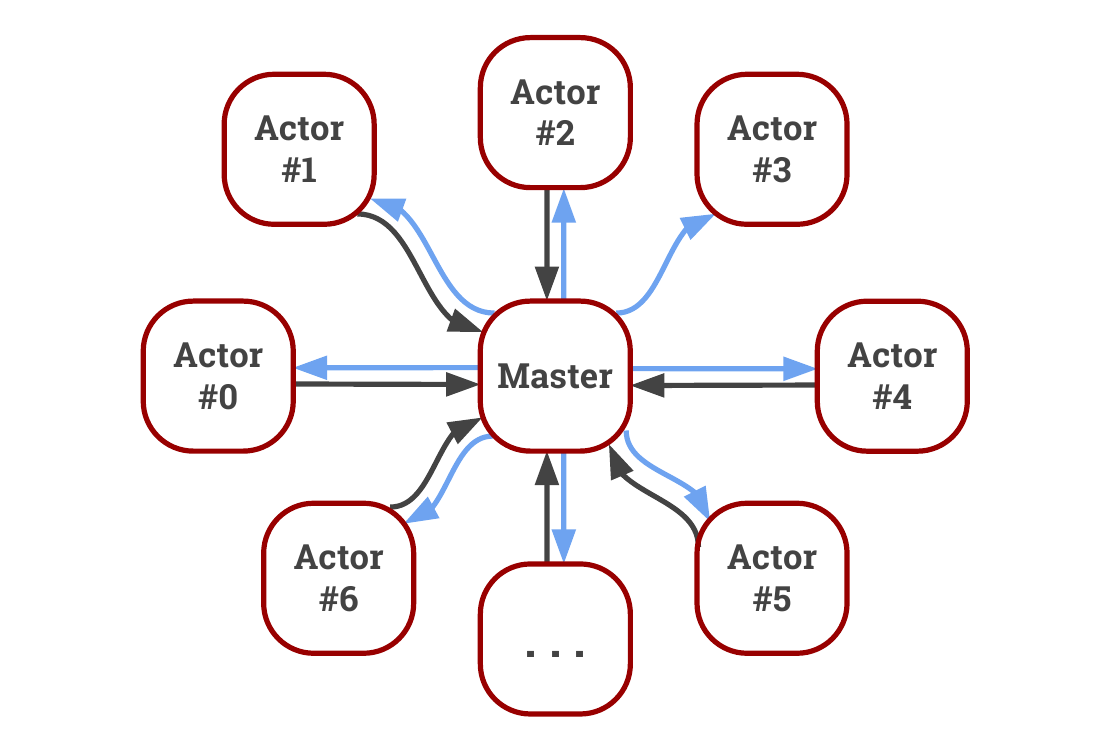}
                \caption{Distributed System Diagram for Evolution Strategies}
                \label{fig:es-diag}
        \end{subfigure}
        \hfill

\caption{Our framework provides abstractions for composing heterogeneous parallel components for different distributed learning algorithms. In this work, we demonstrate two types of algorithms: a) System diagram for on-policy and off-policy reinforcement learning methods. Actor nodes share their experiences with a buffer server, that relays them to the learner server. The updated parameters are broadcast back to the actors via the parameter server. b) System diagram for Evolution Strategies implementation. The actors communicate their experiences to a master node to be broadcasted to all nodes. Parameter updates happen simultaneously in all actors.}

\label{fig:abstraction}
\end{figure*}

\subsection{Algorithm: \surreal}

The algorithm layer at the top of hierarchy uses high-level abstractions of orchestration or communication mechanisms. The clean decoupling between algorithms and infrastructure fulfills our promise to deliver a researcher-friendly and performant open-source framework.


\surreal provides a flexible framework for composing heterogenous parallel components into various types of distributed RL methods. It allows the end users to rapidly develop new algorithms while encapsulating the underlying parallelism. To showcase the flexibility, we develop two distributed learning algorithms with contrastive patterns of parallelism as illustrated in Fig.~\ref{fig:abstraction}: Proximal Policy Optimization (PPO) and Evolution Strategies (ES). PPO requires the coordination of different types of components, including \emph{actor}s, \emph{learner}, \emph{buffer}, and \emph{parameter server}. In contrast, ES is easily parallelizable with an array of homogeneous actors and a master node for model update. We provide an overview of the PPO and ES algorithms in Sec.~\ref{sec:ppo-description} and Sec.~\ref{sec:es-description} respectively. We perform system analysis and quantitative evaluation on our implementations in Sec.~\ref{sec:benchmark} and Sec.~\ref{sec:exp}.

In addition to \sppo and \ses, our prior work~\cite{fan2018surreal} has also implemented a variant of Deep Deterministic Policy Gradient (DDPG)~\cite{lillicrap2015continuous} in the \surreal framework. \sddpg uses the same setup as \sppo, involving actor, learner, replay and parameter server. In contrast to PPO, DDPG is off-policy and reuses observation. Its replay servers thus contain a big replay memory, entries in which are sampled by the learner. From our experiments, we have seen a better learning scalability of \sppo, so we focus on discussing the learning performances of \sppo. A more complete comparisons between these two algorithms can be found in~\cite{fan2018surreal}. \sddpg will be relevant to us in Sec.~\ref{sec:benchmark} where we discuss several design choices to build a system that can accommodate both PPO and DDPG algorithms in a unified abstraction.

\subsubsection{Proximal Policy Optimization}
\label{sec:ppo-description}
Policy gradient algorithms \cite{sutton1998reinforcement} are among the most robust methods for continuous control.
They aim to directly maximize the expected sum of rewards $J(\theta) = \EE_{\tau_\theta}\left[ \sum_t \gamma^{t-1} r(s_t,a_t) \right]$ with respect to the parameters $\theta$ of the stochastic policy $\pi_\theta(a|s)$. The expectation is taken over $\tau_\theta$, which denotes the trajectories induced by $\pi_\theta$ interacting with the environment.
The vanilla policy gradient estimator is given by $\nabla_\theta J_{\mathrm{PG}} = \EE_{\tau_\theta}\left[ \sum_t \nabla_\theta \log \pi_\theta(a_t|s_t) A_t \right]$.
$A_t$ is the advantage function, typically formulated as subtracting a value function baseline from the cumulative reward, $R_t - V(s_t)$.

Policy gradient estimates can have high variance. One effective remedy is to use a \emph{trust region} to constrain the extent to which any update is allowed to change the policy.
Trust Region Policy Optimization (TRPO)~\cite{Schulman2015Trust} is one such approach that enforces a hard constraint on the Kullback-Leibler (KL) divergence between the old and new policies.
It optimizes a surrogate loss
$J_{\mathrm{TRPO}}(\theta) = \EE_{\tau_{\theta_{\mathrm{old}}}}\left[ \sum_t \frac{\pi_\theta(a_t | s_t)}{\pi_{\theta_{\mathrm{old}}}(a_t|s_t)} A_t \right]$  subject to $\mathrm{KL}\left[\pi_{\theta_{\mathrm{old}}} | \pi_\theta\right] < \delta$. More recently, Proximal Policy Optimization (PPO) has been proposed as a simple and scalable approximation to TRPO~\citep{schulman2017proximal}. PPO only relies on first-order gradients and can be easily combined with recurrent neural networks (RNN) in a distributed setting. PPO implements an approximate trust region via a KL-divergence regularization term, the strength of which is adjusted dynamically depending on actual change in the policy in past iterations.
PPO optimizes an alternative surrogate loss
$J_{\mathrm{PPO}}(\theta) = \EE_{\tau_{\theta_\mathrm{old}}}\lb \sum_t \frac{\pi_\theta(a_t | s_t)}{\pi_{\theta_\mathrm{old}}(a_t|s_t)} A_t - \lambda \cdot \mathrm{KL}[\pi_\mathrm{old}|\pi_{\theta}] \rb$, where $\lambda$ is adjusted if the actual KL falls out of a target range. Our \sppo implementation is a distributed variant of the PPO algorithm with the adaptive KL-penalty~\cite{heess2017emergence}.



\subsubsection{Evolution Strategies}
\label{sec:es-description}
Evolution Strategies (ES)~\cite{salimans2017evolution} seeks to directly optimize the average sum of rewards. ES belongs to the family of Natural Evolution Strategies (NES)~\cite{wierstra2008natural} and draws on intuition of natural evolution: a population of model parameters are maintained; at each iteration, each set of parameters are perturbed and evaluated; and models with the highest episodic return is recombined to be the base parameters of next iteration.

Specifically, we let $\theta$ be our model parameters with population distribution $p_\psi(\theta)$ parametrized by $\psi$. We can then take gradient step on $\psi$ to optimize $J(\theta)$: $\nabla_\psi\EE_{\theta\sim p_\psi}J(\theta) = \EE_{\theta\sim p_\psi}\{J(\theta)\nabla_\psi \log p_\psi(\theta)\}$. Following recent work~\cite{salimans2017evolution}, we can reparametrize our population with mean parameter $\theta$ and Gaussian noise $\epsilon\sim N(0,I)$ scaled by $\sigma$ as perturbations to the mean parameters. It allows us to estimate gradient of the return as follows:
$$\nabla_\theta\EE_{\epsilon\sim N(0, I)}J(\theta + \sigma\epsilon) = \frac{1}{\sigma}\EE_{\epsilon\sim N(0, I)}\{J(\theta + \sigma\epsilon)\epsilon\}$$

ES is intrinsically different from PPO. We highlight two major distinctions: First, as shown in the equation above, no gradient is computed for our step estimate. Hence, specialized hardware like GPUs has a less impact on ES than PPO. Second, as suggested by recent work~\cite{lehman2018more}, ES does not seek to optimize each and every actor's performance as in PPO. Instead, ES optimizes the average rewards of the entire population, thereby yielding robust policies that are less sensitive to perturbations.

\lstset{basicstyle=\ttfamily\tiny,breaklines=true}
\begin{figure*}
\setlength{\tabcolsep}{2pt}
\begin{subfigure}{0.31\textwidth}
	\begin{lstlisting}[language=Bash, linewidth=\textwidth]
>> python cloudwise-gke.py
Please give your cluster a name
> kuflexes
Do you wish to create a node pool with 32 CPUs and 1 Nvidia V100 GPU per machine?
> Yes
...
Generating kuflexes.tf.json
Use "terraform apply" to create the cluster
\end{lstlisting}
	\caption{Interacting with \cloudwise CLI to generate cluster specification. }
	\label{fig:code-cloudwise}
\end{subfigure}
\hfill
\begin{subfigure}{0.31\textwidth}
	\begin{lstlisting}[language=Python, linewidth=\textwidth]
# launch_experiment.py
exp = symphony.Experiment()
learner = exp.new_process(cmd="python learner.py")
replay = exp.new_process(cmd="python replay.py")
replay.binds("replay")
learner.connects("replay")
cluster.launch(exp)
...
\end{lstlisting}
	\caption{Declaring a distributed experiment using \symphony. Network connnections are defined in a straightforward way.}
	\label{fig:code-symphony-declaration}
\end{subfigure}
\hfill
\begin{subfigure}{0.31\textwidth}
	\begin{lstlisting}[language=Python, linewidth=\textwidth]
# learner.py
host = os.environ["SYMPH_REPLAY_HOST"]
port = os.environ["SYMPH_REPLAY_PORT"]
sender = caraml.Sender(host=host, port=port, serializer=pickle.dumps)
sender.send(data)
# replay.py
...
receiver = caraml.Receiver(host=host, port=port, serializer=pickle.loads)
data = receiver.recv()
\end{lstlisting}
	\caption{\symphony maps network address to environment variables so network connection is compatible across platforms.}
	\label{fig:code-symphony-usage}
\end{subfigure}
\\
\begin{subfigure}{0.31\textwidth}
	\begin{lstlisting}[language=Python, linewidth=\textwidth]
# launch_experiment.py
exp = symphony.Experiment()
learner = exp.new_process(cmd="python learner.py")
dispatcher = symphony.Dispathcer(cluster= "kuflexes.tf.json")
dispatcher.assign_to_nodepool(learner,
     "v100-nodepool", cpu=5, gpu=1)
	\end{lstlisting}
	\caption{\symphony provides scheduling bindings to underlying clusters.}
	\label{fig:code-scheduling}
\end{subfigure}
% ...
\hfill
\begin{subfigure}{0.31\textwidth}
	\begin{lstlisting}[language=Bash, linewidth=\textwidth]
# monitoring experiments
>> kuflexes list-experiments
Humanoid
Cheetah
>> kuflexes switch-experiment Cheetah
>> kuflexes list-processes
actor-0
...
>> kuflexes logs actor-0
actor-0 running ...
\end{lstlisting}
	\caption{\symphony provides experiment management features to view experiments and processes.}
	\label{fig:code-exp-management}
\end{subfigure}
\hfill
\begin{subfigure}{0.31\textwidth}
	\begin{lstlisting}[language=Python, linewidth=\textwidth]
# ppo_better.py
class BetterPPOLearner(PPOLearner):
	...

# launch_ppo_better.py
from ppo_better import BetterPPOLearner
launcher = surreal.PPOLauncher(learner=BetterPPOLearner)
launcher.main()
	\end{lstlisting}
	\caption{\surreal is designed to be easily extensible. Each component can be subclassed to show custom behaviors.}
	\label{fig:code-extension}
\end{subfigure}
\caption{Code snippets for various \surreal use cases. The \surrealsys stack provides support for cluster creation, experiment declaration, network communication, resource allocation, experiment management and extensions.}
\label{fig:code}
\end{figure*}


\section{Using \surreal as a Researcher}
\label{sec:surreal-infras}

The \surrealsys stack facilitates the development and deployment of distributed learning algorithms from the ground up. In this section, we provide a walk-through of our system implementation from an end-user's perspective. We will start with a bare-bone cloud account, walk up the ladders of abstractions, and reach the end goal of implementing distributed RL algorithms.

\subsection{Provision a Computing Cluster}
\surreal reaches its full potential when running on a Kubernetes cluster. \cloudwise is a helper for setting up a Kubernetes cluster suitable for RL runtime. As shown in Fig.~\ref{fig:code-cloudwise}, \cloudwise allows intuitive user interactions and generates a terraform definition which can be used to boot up the cluster. We also provide terraform files defining the cluster used in our experiments.

\subsection{Defining a Distributed Experiment}
\symphony provides a simple way of declaring a distributed environment. Each distributed experiment is made up of multiple processes. They are declared in the launch script. Network communication patterns are specified by letting each process declare the services it provides and the services it binds to. An example can be seen in Fig.~\ref{fig:code-symphony-declaration}.

Thanks to these platform-agnostic communication patterns, launch mechanisms in \symphony can automatically configure different platforms to expose the necessary network interfaces. For example, \symphony would create and configure platform-specific ``services'' when Kubernetes or Docker compose is used. It will assign local port numbers in Tmux mode. These arranged addresses are provided in the form of environment variables. See Fig.~\ref{fig:code-symphony-usage} for an example. This design allows smooth transition from local, small scale development to cloud-based large-scale deployments. 

\subsection{Scheduling Processes in a Flexible Way}
\label{sec:scheduling-flexibility}
\symphony provides bindings with various platforms to enable flexible scheduling (see Fig.~\ref{fig:code-scheduling}). Allowing processes to claim sufficient amounts of resources ensures that components run at full speed. Being able to control how much resource to allocate can drastically improve efficiency of algorithms. For instance, we demonstrate in Sec.~\ref{sec:batching-actors} that running multiple actors on the same GPU can substantially improve the resource utility of a learning algorithm.

\subsection{Dockerizing for Reproducibility}
\surreal experiments on Kubernetes are always dockerized to ensure scalability. This guarantees good reproducibility as the code for every launched experiment resides in a specific docker image in the registry. Moreover, \symphony provides serialization for experiment declaration, saving not only source code but also the launching scripts. To ease the process of building docker images, \symphony provides docker building utilities. For example, one can assemble files from multiple locations and build them into a single docker image. 

\subsection{Managing Experiments Conveniently}
\symphony also provides utilities to manage multiple experiments running on the same cluster. An example can be seen in Fig.~\ref{fig:code-exp-management}. One can view all running experiments, view all running processes of an experiment, and view logs of each process. These functionalities also allow team members to cooperate and exchange progress.

\begin{figure}
\centering
\begin{subfigure}[b]{0.49\linewidth}
    \includegraphics[width=1.\textwidth]{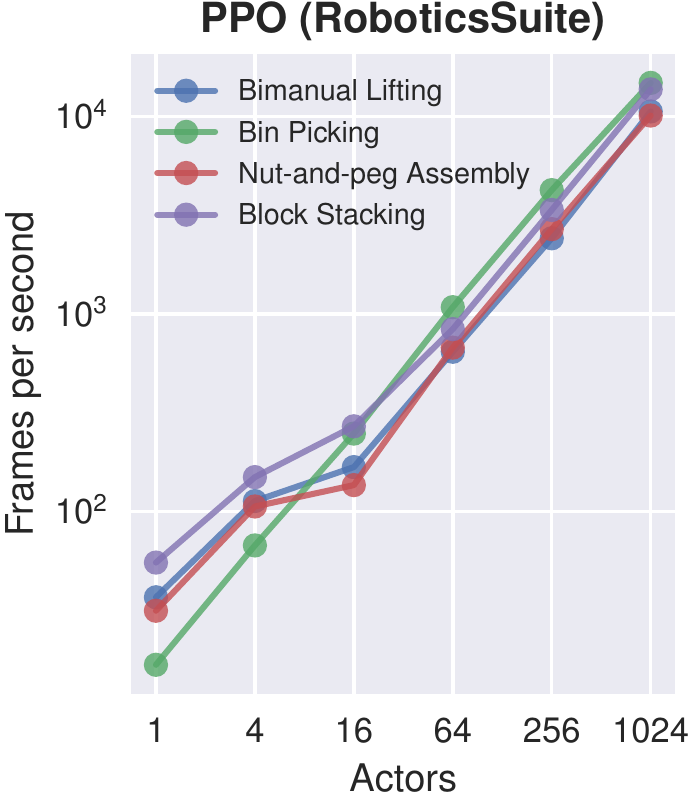}
\end{subfigure}
\begin{subfigure}[b]{0.49\linewidth}
    \includegraphics[width=1.\textwidth]{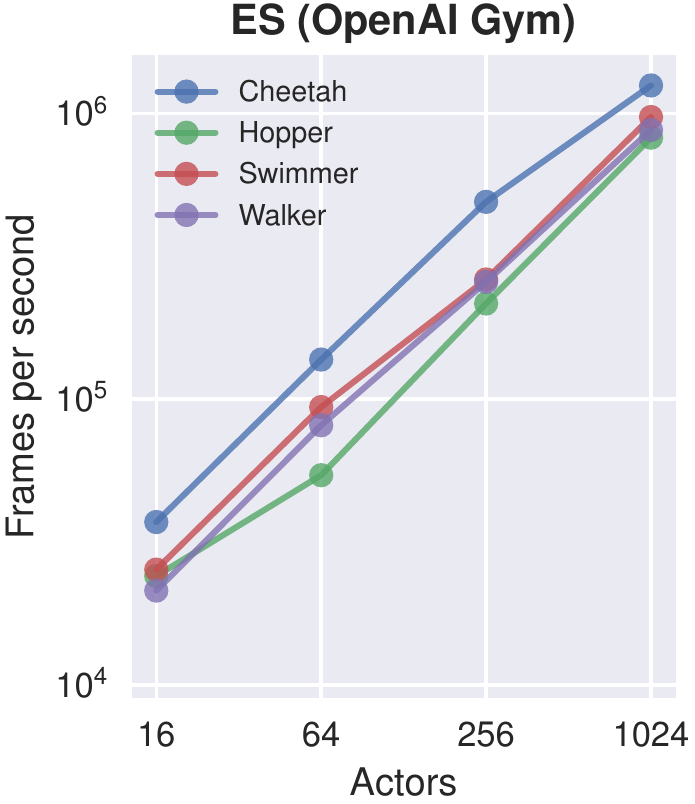}
\end{subfigure}
    \vspace{-4mm}
    \caption{Scalability of environment interactions (in FPS) with respect to the number of actors on \sppo (Robotics Suite) and \ses (OpenAI Gym).}
    \label{fig:actor-throughput}
    \vspace{-4mm}
\end{figure}

\subsection{Running \surreal Algorithms}
\surreal algorithms are developed with tools provided by \cloudwise, \caraml, and \symphony. \surreal is designed to be easily extensible. One can implement a new \texttt{Learner} subclass with different model architectures and different learning schemes. The \texttt{Launcher} class provides a unified interface such that custom-built \surreal components can be properly executed by scheduling code (see Fig.~\ref{fig:code-extension}).


%

\section{Evaluation Tasks}

To examine the quality of our implementation of \surrealsys, we examine the performance and efficiency of \sppo and \ses in solving challenging continuous control tasks with these two algorithms.

\begin{figure}
    \centering
\begin{subfigure}[b]{.48\linewidth}
    \includegraphics[width=1.0\linewidth]{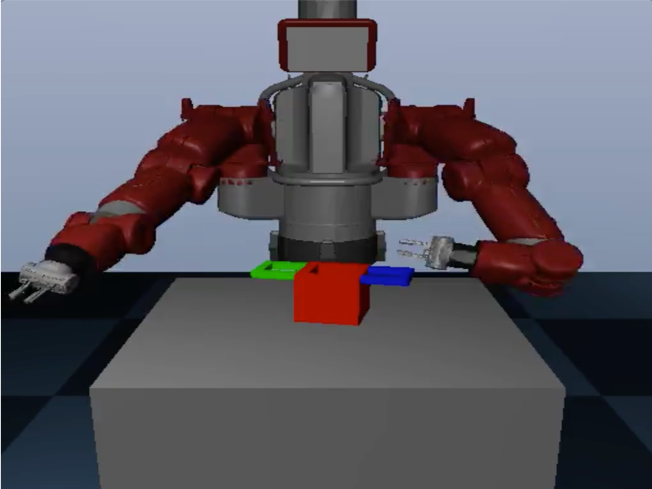}
    \caption{Bimanual Lifting}
    \label{fig:baxter_lifting_img}
\end{subfigure}
\begin{subfigure}[b]{.48\linewidth}
    \centering
    \includegraphics[width=1.0\linewidth]{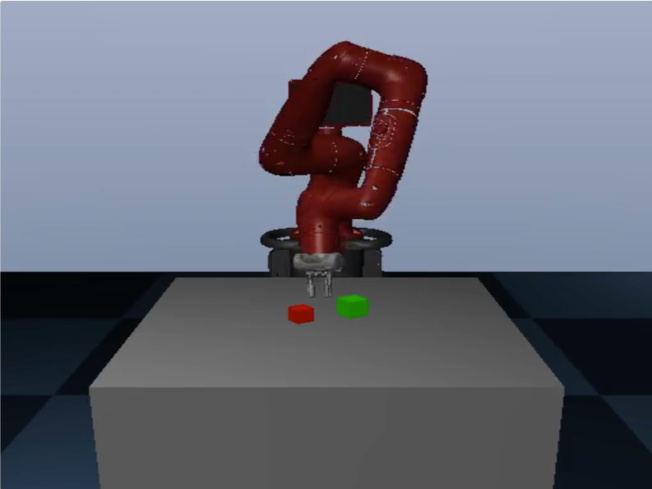}
    \caption{Block Stacking}
    \label{fig:block_stacking_img}
\end{subfigure}
\begin{subfigure}[b]{.48\linewidth}
    \centering
    \includegraphics[width=1.0\linewidth]{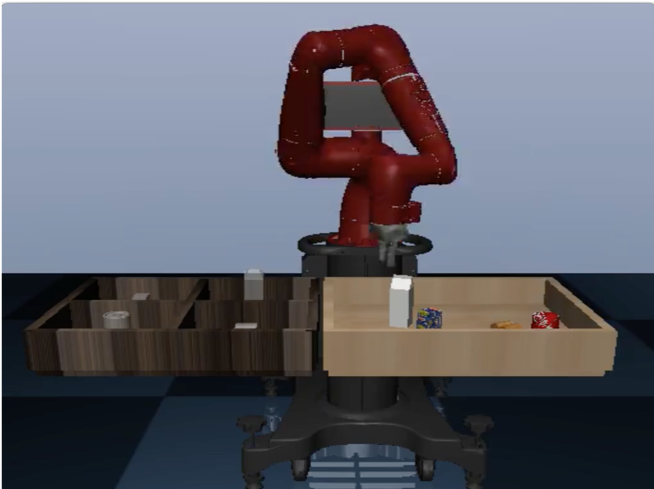}
    \caption{Bin Picking}
    \label{fig:bin_picking_img}
\end{subfigure}
\begin{subfigure}[b]{.48\linewidth}
    \centering
    \includegraphics[width=1.0\linewidth]{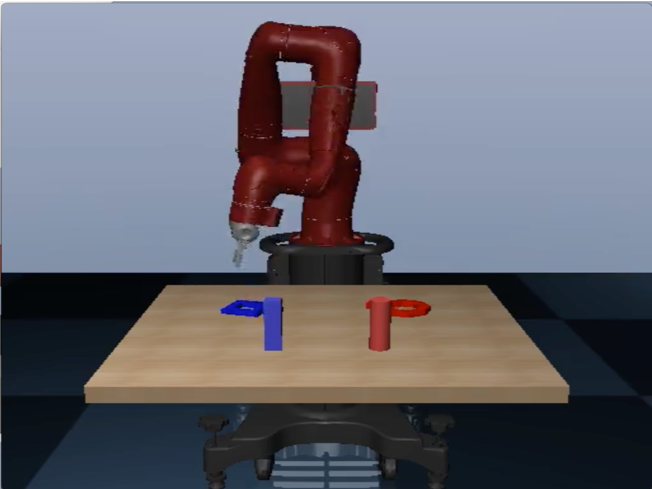}
    \caption{Nut-and-peg Assembly}
    \label{fig:nut_and_peg_assembly_img}
\end{subfigure}
    \caption{Screenshots of our robot manipulation tasks in Robotics Suite~\cite{fan2018surreal}:
    \textbf{a) Bimanual Lifting.} The goal is to grab the pot by both handles and lift it off the table;
    \textbf{b) Block Stacking.} The robot picks up the red block and places it on top of the green one;
    \textbf{c) Bin Picking.} The robot picks up each item and place each into its corresponding bin;
    \textbf{d) Nut-and-peg Assembly.} The goal is to place the nut over and around the corresponding pegs.}
    \label{fig:env-list}
    \vspace{-3mm}
\end{figure}

We evaluate our \sppo implementation on four robot manipulation tasks in Robotics Suite~\cite{fan2018surreal}. These tasks consist of tabletop manipulation tasks with single-arm and bimanual robots. We provide screenshots of these manipulation tasks in Fig.~\ref{fig:env-list}. All four tasks are complex and multi-stage, posing a significant challenge for exploration. In particular, we evaluate the strengths of \sppo in learning visuomotor policies from pixel to control. The neural network policy takes as input RGB images and proprioceptive features and produces joint velocity commands at 10Hz. We evaluate our \ses implementation on locomotion tasks in OpenAI Gym~\cite{brockman2016openai}. We report quantitative results on four locomotion tasks, including HalfCheetah, Hopper, Swimmer, and Walker2d. These standard tasks are used by prior work~\cite{salimans2017evolution} for benchmarking Evolution Strategies implementations.

\section{Systems Benchmarking}
\label{sec:benchmark}

Built to facilitate large distributed computation, \surrealsys allows algorithms to utilize large amounts of computing power. In this section, we investigate the system scalability with the \surreal learning algorithms.

\subsection{System Scalability}
\begin{figure}
\centering
\begin{subfigure}[b]{0.49\linewidth}
	\centering
	\includegraphics[trim={10px 5px 15px 10px},clip,width=1.0\linewidth]{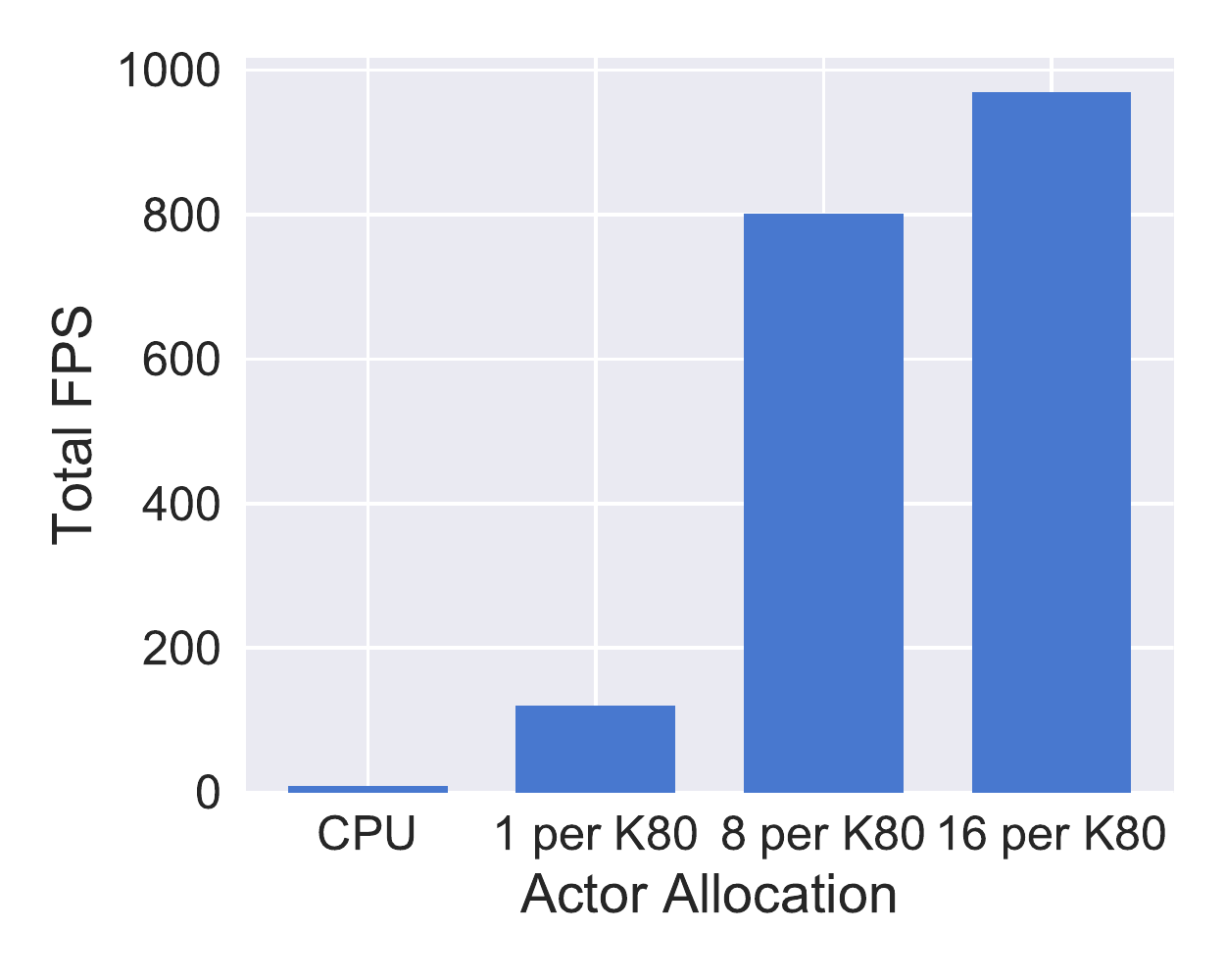}
	\caption{Total throughput}
	\label{fig:total-throughput}
\end{subfigure}
\hfill
\begin{subfigure}[b]{0.49\linewidth}
	\centering
	\includegraphics[trim={10px 5px 15px 10px},clip,width=1.0\linewidth]{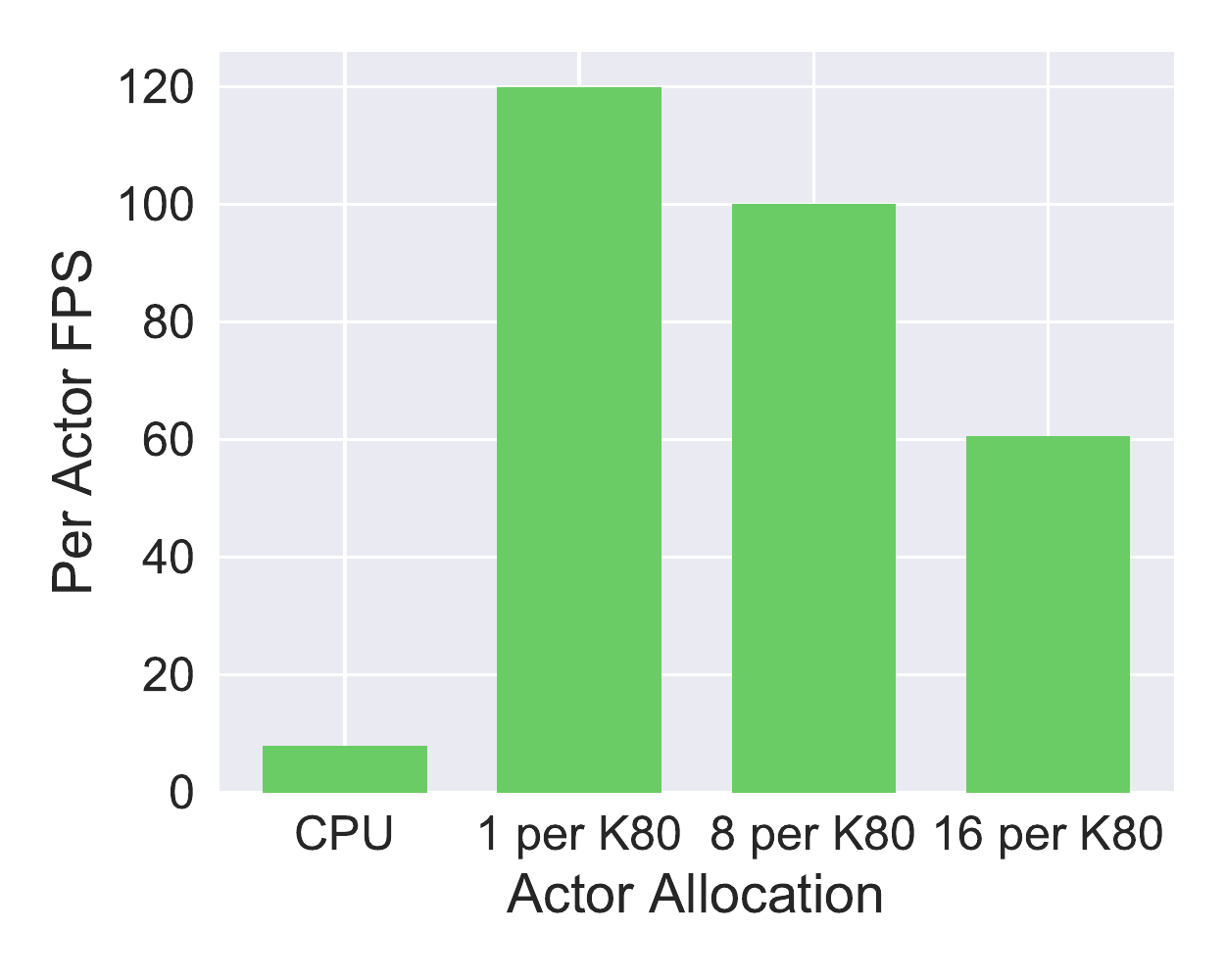}
	\caption{Throughput per actor}
	\label{fig:actor-latency}
\end{subfigure}
\vspace{-2mm}
\caption{Speed of policy evaluation on the Block Lifting task with pixel observations. We compare system speed with and without GPU acceleration. In the cases where a NVIDIA K80 GPU is used, we consider sharing the GPU among multiple actors. \textbf{a) Total throughput.} The number of environment frames collected by all actors. \textbf{b) Throughput per actor.} The number of environment frames collected by a single actor.}
\label{fig:batch-ablation}
\end{figure}

We run our experiments on Google Cloud Kubernetes Engine. \sppo actors for the Robotics Suite tasks were trained with 1 learner on a machine with 8 CPUs and a Nvidia V100 GPU, accompanied by 1, 4, 16, 64, 256, or 1024 actors. Multiple actors were run on the same machine in order to fully utilize the GPU (see Sec.~\ref{fig:batch-ablation} for details); for tasks Block Stacking, Nut-and-peg Assembly, and Bimanual Lifting, we place 16 actors on each machine of 8 CPUs and 1 Nvidia P100 GPU. For the Bin Picking task, we place 8 actors on each machine due to memory constraints. For OpenAI Gym were the actor nodes do not need a GPU for rendering and hence we use a dedicated 2-CPU machine per actor.
The \ses experiments were run on CPU only machines with 32 cores each with up to 32 actors batched per machine. 

We measure \sppo's scalability in terms of total actor throughput, which is the total environment frames generated by all actors combined per second. Fig.~\ref{fig:actor-throughput} shows our total actor throughput on the Block Stacking task. We see an approximately linear increase in total frames generated with an increase in actors. 


We measure \ses's scalability in terms of total environments interactions per second. 
Fig.~\ref{fig:actor-throughput} shows an almost linear scaling of the same with the number of actors on various Gym environments.

\begin{table}
	\centering
	\begin{tabular}{c|c}
	 & Throughput \\
	\# Replay Shards & ($\times10^3$ observations/s) \\ \hline
	1 & 3.5 \\
	3 & 4.5 \\
	5 & 4.5 \\
	\hline
	\end{tabular}
	\caption{Effect of sharding the replay buffer when using \sddpg with 128 actors. Throughput measures the total number of actor observations received by the replay server.}
	\label{table:sharding-replay}
\end{table}

\subsection{Load Balanced Replay}
\label{sec:sharding-replays}
As described in Sec.~\ref{sec:symphony-scalability}, the combination of \symphony, \caraml, and Kubernetes enables simple and effective horizontal scalability. One example is sharding replay when training \sddpg on the Gym Cheetah environment. Table~\ref{table:sharding-replay} shows the number of experiences handled in total with 1, 3, and 5 sharded replays in presence of 128 actors. A single replay buffer can no longer handle all the actor outputs, becoming the bottleneck of the system. Three load-balanced replay buffers resolves congestion. More replays does not further improve overall throughput.

\subsection{Batching Actors}
\label{sec:batching-actors}
As described in Sec.~\ref{sec:scheduling-flexibility}, \surrealsys supports a flexible scheduling scheme. It allows us to utilize resources more efficiently. We demonstrate this point by using \symphony to batch multiple \sppo actors on the same GPU. Fig.~\ref{fig:batch-ablation} shows policy evaluation speed on vision-based Block Lifting tasks, measured by environment frames per second. Each actor gets $1$, $\frac{1}{8}$, or $\frac{1}{16}$ of a NVIDIA K80 GPU. Actor speed on CPU is provided for reference.
Fig.~\ref{fig:total-throughput} shows per GPU \textit{throughput} measured by the total number of frames generated by all actors on the same GPU. Higher throughput means that experiments are more resource-efficient. Sharing a GPU among multiple actors achieves a better throughput and thus increases GPU utilization. However, as seen in Fig.~\ref{fig:actor-latency}, each actor's speed decreases with an increasing number of actors sharing the same GPU. We set 16 actors per GPU in our experiments to attain the best trade-off between number of actors and per-actor throughput.

\subsection{Serialization and Communication}
\label{sec:caraml-speed-ablation}
In distributed RL algorithms, feeding large amounts of training data to the learner requires fast network communication and fast serialization. Doing everything on the main Python process would bottleneck the entire system. \caraml offloads communication to separate processes to circumvent global interpreter lock. It then serializes data and saves them to a shared memory to minimize interprocess data transfer. PyArrow is used to serialize data because it provides fast deserialization. The speed-up of these optimizations is measured on reinforcement learning algorithms \sddpg and \sppo trained on the Gym Cheetah environment. Speed-up measured by learner iterations per second is reported in Table.~\ref{table:caraml-effect}. The speed-up in DDPG is more pronounced as PPO is less communication bound than DDPG. In comparison, the PPO learner iteration performs more computation than that of DDPG.

\begin{table}
\centering
\begin{tabular}{l|cc}
 & PPO (iters/s) & DDPG (iters/s) \\ \hline
without optimization & 5.1 & 5.4 \\
with optimization & \textbf{38} & \textbf{54} \\ \hline
\end{tabular}
\caption{Learner iteration speed for PPO and DDPG with and without the communication optimizations of \caraml. Numbers are obtained when training on the Gym Cheetah environment.}
\label{table:caraml-effect}
\end{table}

\section{Quantitative Evaluation}
\label{sec:exp}

Here we further examine the effectiveness and scalability of \surrealsys in learning efficiency and agent performance. To this end, we implement the Proximal Policy Optimization (PPO) and Evolution Strategies (ES) algorithms using the APIs provided by our framework. These two algorithms have distinct characteristics: ES is embarrassingly parallel with a large number of homogenous actors, while PPO requires the coordination between heterogeneous types of parallel components. Our primary goal is to answer the following two questions: 1) are our implementations of these two algorithms capable of solving challenging continuous control tasks and achieve higher performances than prior implementations, and 2) how well can our distributed algorithms scale up with an increasing amount of computational resources? The experiment setup used here is same as the one described in Sec.~\ref{sec:benchmark}.




\subsection{\sppo Evaluation}

For each of the Robotics Suite tasks, we train a PPO model which takes as input an $83\times83\times3$ RGB image and proprioceptive features (e.g., arm joint positions and velocities). The image is passed through a convolutional encoder. The resulting activations are flattened to a vector and concatenated to the proprioceptive features, which is further fed through an LSTM layer of 100 hidden units. The output of the LSTM layer is passed through additional fully-connected layers of the actor network and the critic network for producing the final outputs.

\begin{figure*}[t!]
\centering
\begin{subfigure}[b]{.245\linewidth}
    \centering
    \includegraphics[width=1.0\linewidth]{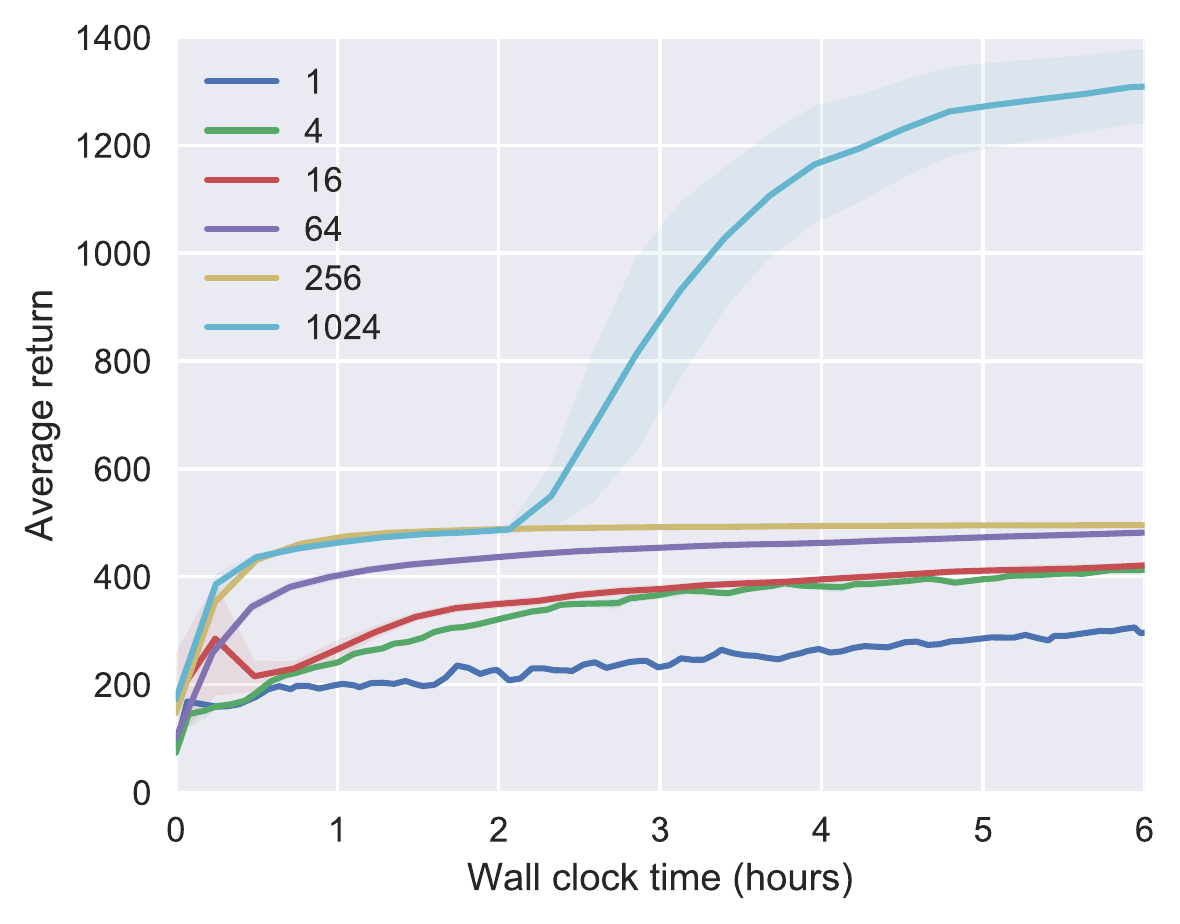}
    \caption{Bimanual Lifting}
    \label{fig:baxter_lifting}
\end{subfigure}
\begin{subfigure}[b]{.245\linewidth}
    \centering
    \includegraphics[width=1.0\linewidth]{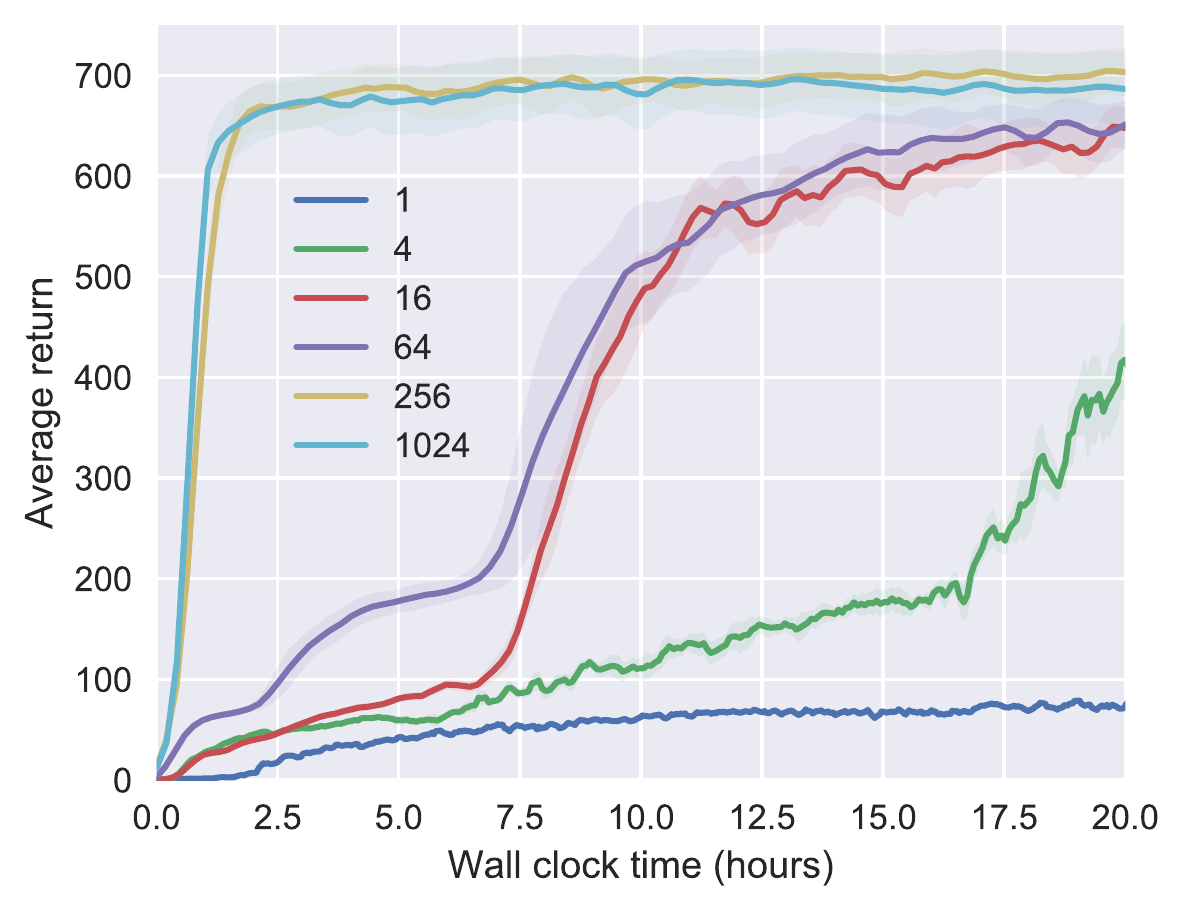}
    \caption{Block Stacking}
    \label{fig:block_stacking}
\end{subfigure}
\begin{subfigure}[b]{.245\linewidth}
    \centering
    \includegraphics[width=.995\linewidth]{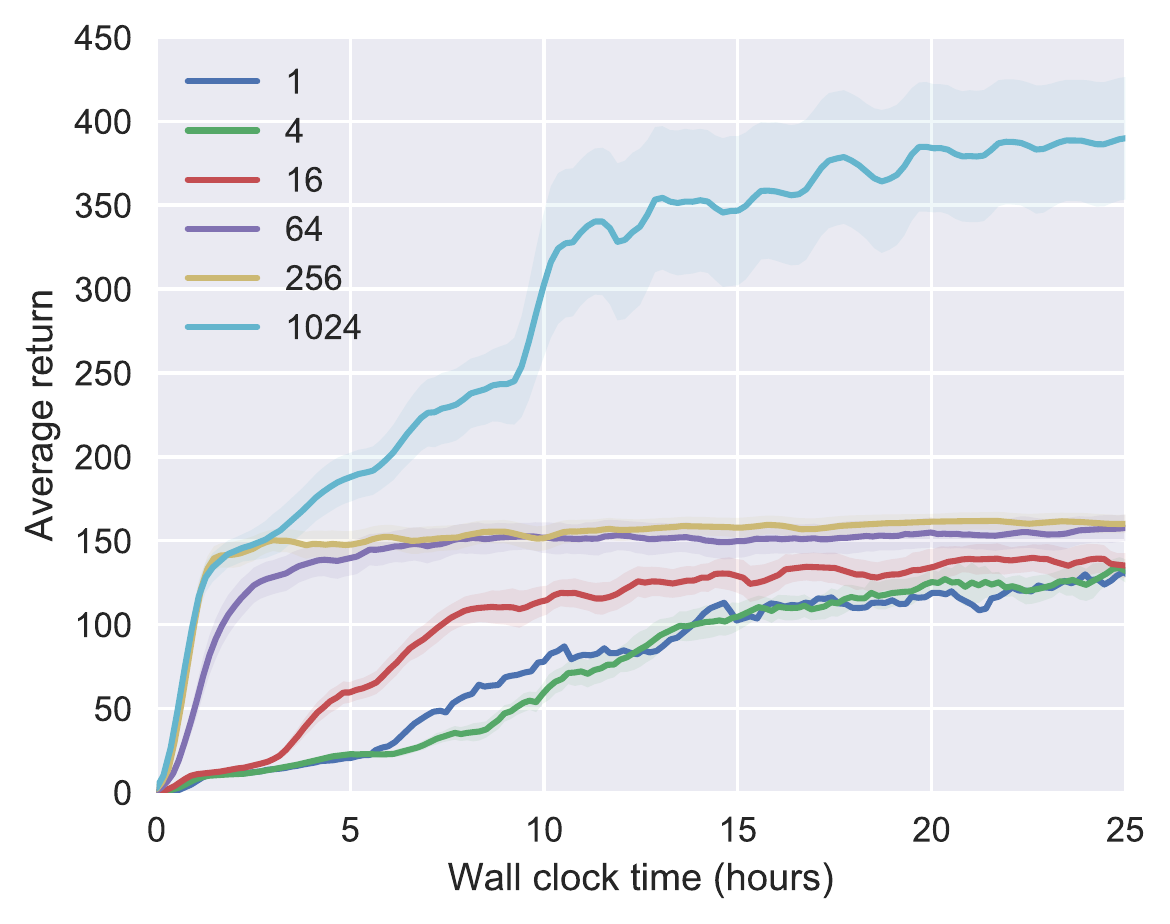}
    \caption{Bin Picking}
    \label{fig:bin_picking}
\end{subfigure}
\begin{subfigure}[b]{.245\linewidth}
    \centering
    \includegraphics[width=1.0\linewidth]{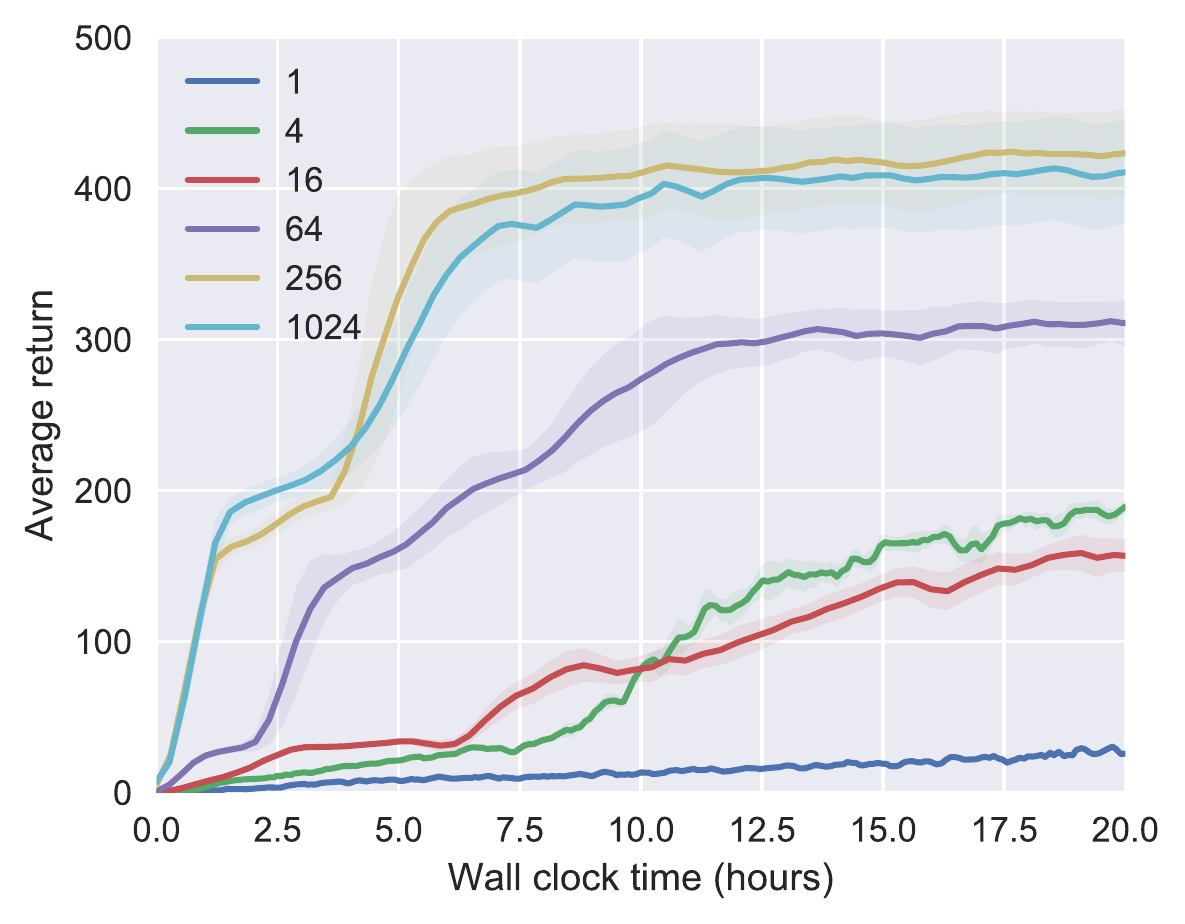}
    \caption{Nut-and-peg Assembly}
    \label{fig:nut_and_peg_assembly}
\end{subfigure}
\vspace{-2mm}
\caption{Scalability for \sppo experiments on Robotics Suite tasks. The solid line represents mean return and the translucent region around the mean represents one standard deviation. PPO models were trained on 1, 4, 16, 64, 256, and 1024 actors.}
\label{fig:robotics-scalability}
\end{figure*}

Fig.~\ref{fig:robotics-scalability} reports results of our PPO implementations with varying number of actors. We see an overall trend towards higher scores and faster training with an increased number of actors. In the Bimanual Lifting and Bin Picking tasks, we observe a substantial benefit of using 1024 actors, where the learned policy is able to advance to later stages of the tasks than the runs with fewer actors. The trained PPO model  is able to pick up the can and place it into the bin in the Bin Picking task, and in the Bimanual Lifting task is able to lift the pot off the table. In the Block Stacking and Nut-and-peg Assembly tasks, the experiments with 256 actors and 1024 actors managed to converge to the same level of performance, which achieve both faster and better learning than the ones with fewer actors.

\subsection{\ses Evaluation}

Our ES implementation takes low-dimension state features as input. These features are passed through a network with two fully-connected layers. The activations are passed through a final fully-connected layer and tanh activation to map to the predefined action space. To interact with the environment, we use a stochastic policy with mean generated from the neural network and fixed standard deviation of 0.01. For the population noise, we sample from Gaussian distribution centered at 0 with standard deviation 0.02. Additionally, we discretize the action space to 10 bins per component for Swimmer and Hopper to aid exploration.

Fig.~\ref{fig:robotics-scalability} shows performances of our ES implementation on various locomotion tasks with increasing number of actors. We see that \ses scales well with a large number of actors, leading to faster convergence and shorter training time. We hypothesize that it is attributed to the effects of improved exploration with more actors executing diverse policies. Through experimentation, we find that the scale of standard deviation of Gaussian noise compared to the scale of parameters greatly impacts learning efficacy. High noise values cause learning to be unstable whereas low noise values cause slow learning.

\begin{figure*}[t!]
\begin{subfigure}[b]{.245\linewidth}
    \centering
    \includegraphics[width=1.0\linewidth]{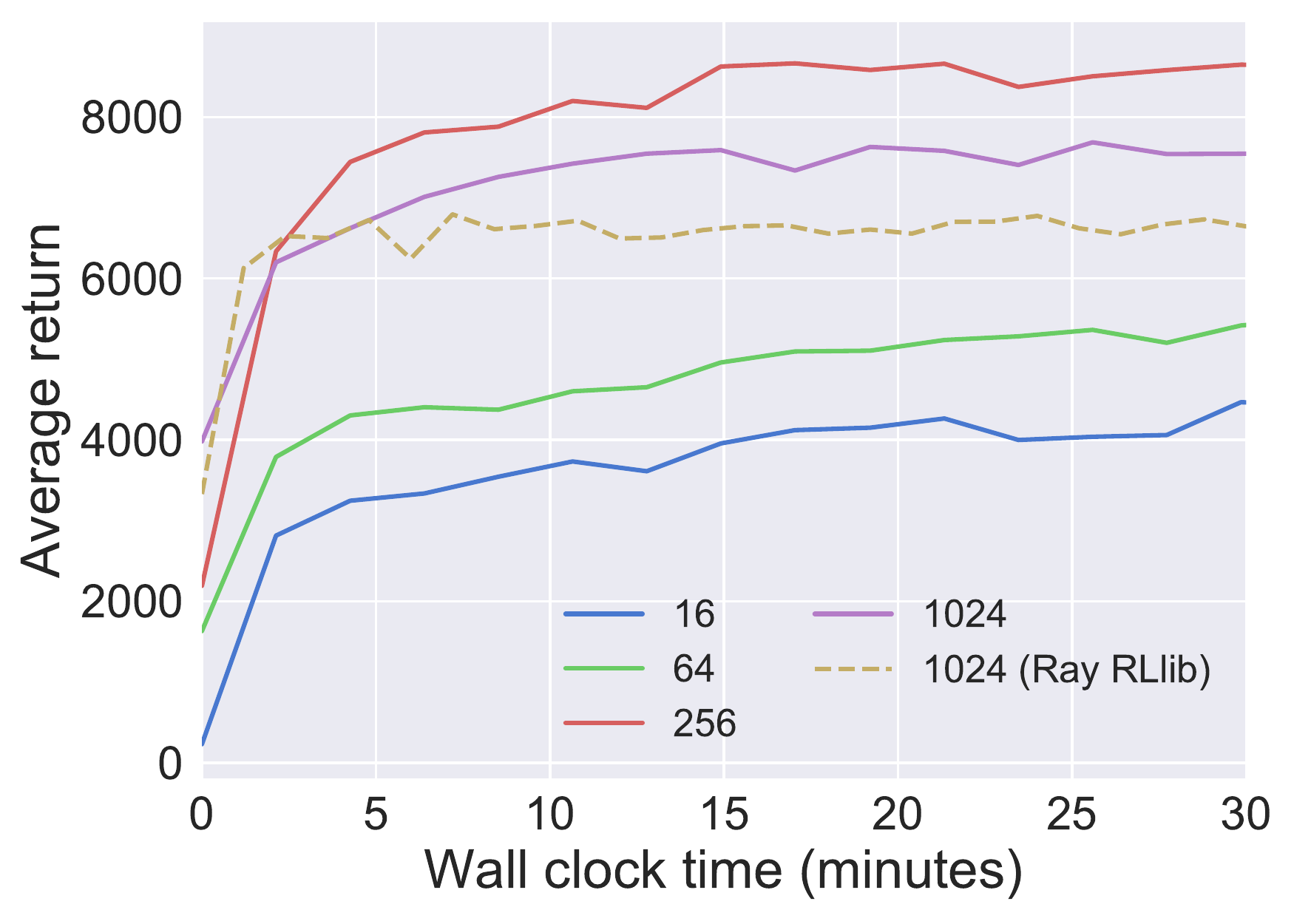}
    \caption{HalfCheetah}
\end{subfigure}
\begin{subfigure}[b]{.245\linewidth}
    \centering
    \includegraphics[width=1.0\linewidth]{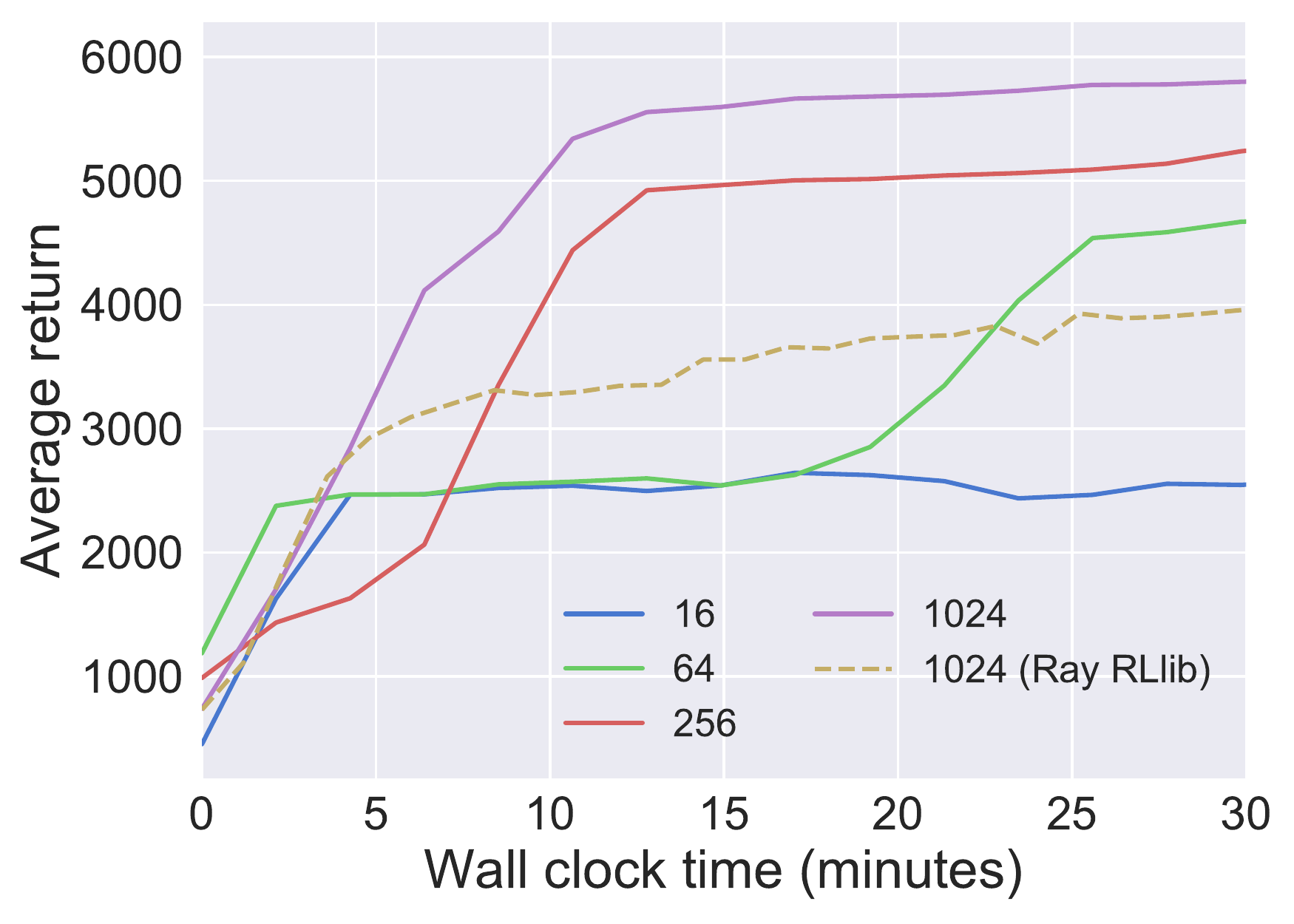}
    \caption{Walker2d}
\end{subfigure}
\begin{subfigure}[b]{.245\linewidth}
    \centering
    \includegraphics[width=1.0\linewidth]{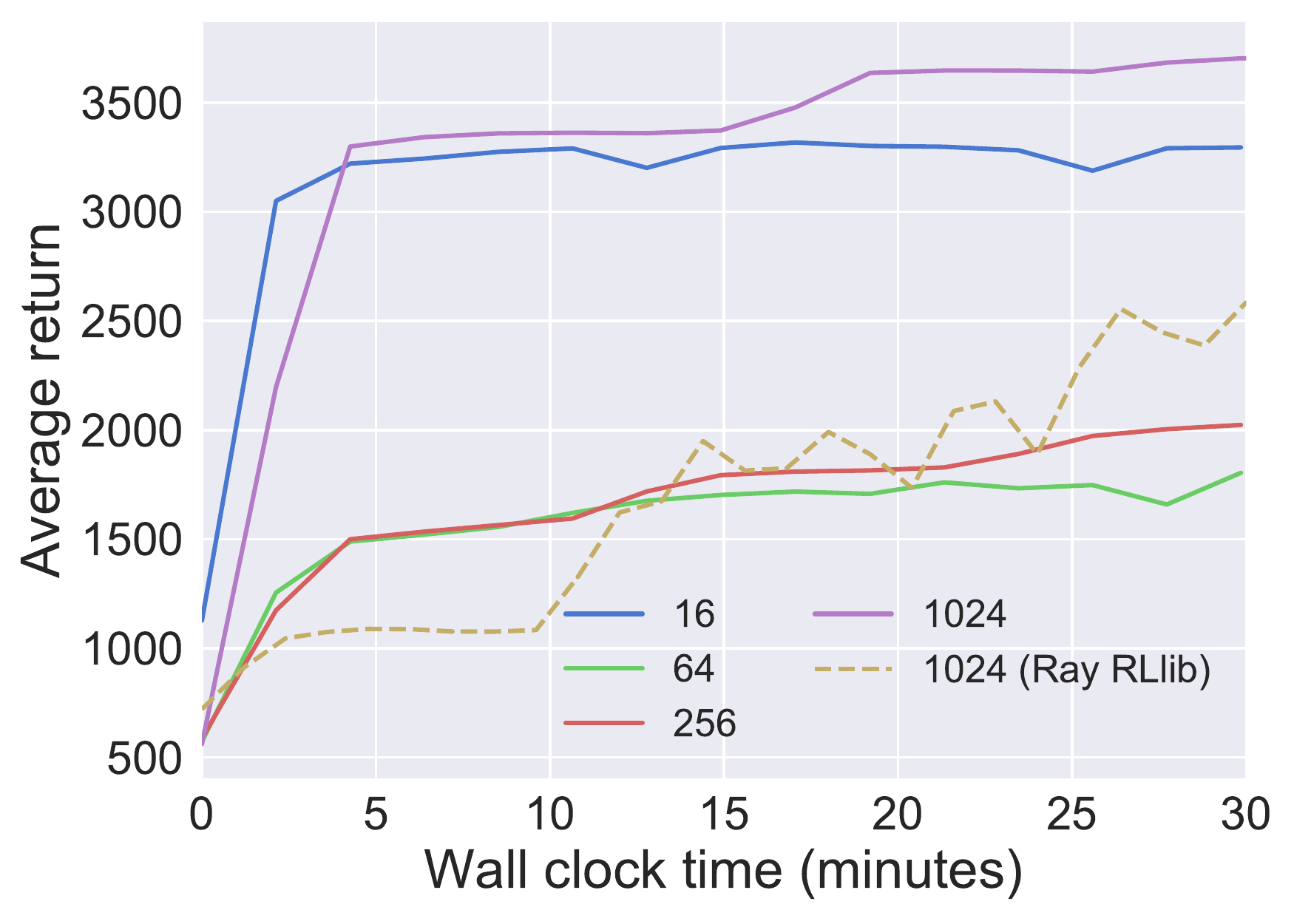}
    \caption{Hopper}
\end{subfigure}
\begin{subfigure}[b]{.245\linewidth}
    \centering
    \includegraphics[width=1.0\linewidth]{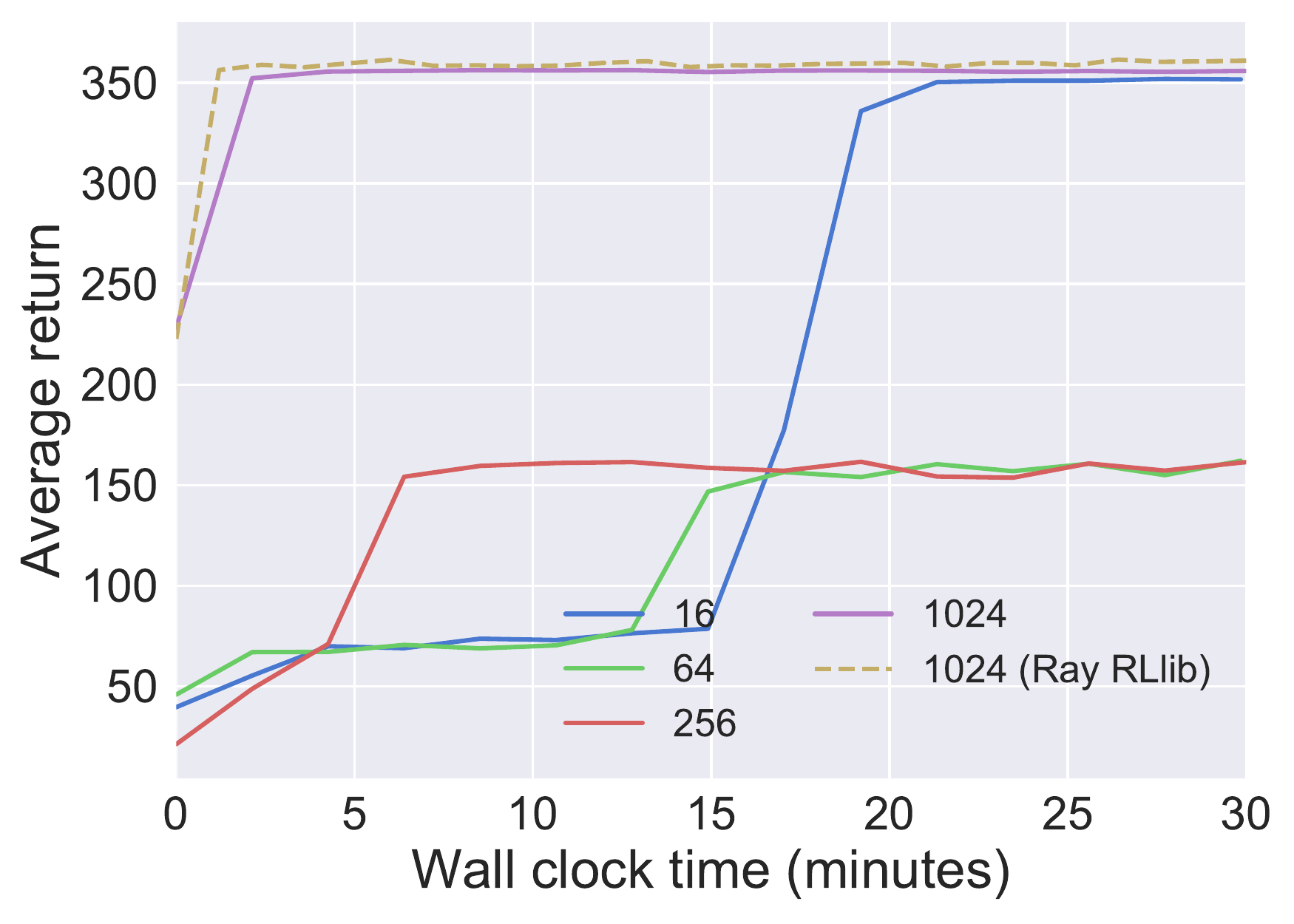}
    \caption{Swimmer}
\end{subfigure}
\vspace{-2mm}
    \caption{Scalability of \ses on OpenAI Gym tasks. The dashed line represents the best curve obtained by Ray RLlib~\cite{liang2017ray} with 1024 actors and the rest correspond to our implementation trained with 16, 64, 256, 1024 actors.}
\label{fig:es_vs_ray}
\end{figure*}

\ses is very competitive with the state-of-the-art reference implementation~\cite{liang2017ray} as shown in Fig~\ref{fig:es_vs_ray}. Across all four Gym environments, \ses outperforms Ray RLlib in terms of both final performance and wall-clock time.

\section{Related Work}

%

Reinforcement learning methods, powered by deep neural networks, often require a significant amount of experience for learning. This has accentuated the advantages of large-scale distributed learning methods for learning efficiency. A series of deep learning models and algorithms have been proposed for large-scale sequential decision making. One notable class of architectures and algorithms scales deep RL methods by using asynchronous SGD to combine gradients and update global parameters. The \emph{Gorila} architecture~\citep{nair2015massively} proposes to use multiple actors, multiple learners, a distributed experience replay memory, and a parameter server that aggregates and applies gradients shared by the learners. A3C~\citep{mnih2016asynchronous} instead utilizes several CPU threads on a single machine, where each thread is an actor-learner that shares gradients with the other threads. A distributed version of PPO was also introduced by \citep{heess2017emergence} with several workers that collect experience and send gradients to a centralized learner. However, sharing gradients instead of experience is less desirable since gradients become stale and outdated much more quickly, especially for off-policy methods. 

Another class of architectures scales deep RL methods via several actors that solely collect and send experience, a distributed replay memory and parameter server, and one centralized learner that performs parameter updates using experience sampled from the replay memory. Ape-X \citep{horgan2018distributed}, IMPALA \citep{espeholt2018impala}, and \sppo implementation fall into this category. In contrast to prior work which focused on off-policy methods, we extend this paradigm to accommodate on-policy learning.

Prior work has also shown the efficacy of distributed algorithms based on evolutionary computation. \citep{salimans2017evolution} applied Evolution Strategies (ES) to common deep RL benchmarks and found that their method could scale to many more distributed workers than RL algorithms while achieving competitive results. Similarly, \citep{such2017deep} found that a simple genetic algorithm (GA) could leverage distributed computation much more effectively than RL methods, which suffer from bottlenecks during learning, and also produce competitive results on several domains.

Existing frameworks of distributed RL algorithms exploit both data parallelism and model parallelism to learn on massive data~\cite{dean2012large}. Open-source libraries for distributed RL include OpenAI Baselines~\citep{baselines}, TensorFlow Agents~\citep{hafner2017tensorflow},  Neuroevolution~\citep{such2017deep}, etc. These libraries focus on algorithm implementations built on third-party learning frameworks, such as Tensorflow and PyTorch, without providing infrastructure support for computing runtime. Ray RLlib~\cite{liang2017ray} is built on the Ray distributed framework designed for machine learning applications. It also has flexibility in supporting different types of distributed algorithms, including Evolution Strategies (ES) and Proximal Policy Optimization (PPO). In contrast to Ray RLlib, \surreal also provides a set of toolkits that sit between the learning algorithms and the hardware for scalable deployment in different computing platforms.

\section{Conclusion}


We introduced \surrealsys, a four-layer distributed learning stack for reproducible, flexible, scalable reinforcement learning. It enables an end-user to deploy large-scale experiments with thousands of CPUs and hundreds of GPUs, facilitating researchers to rapidly develop new distributed RL algorithms while reducing the effort for configuring and managing the underlying computing platforms. We have released the complete source code of our \surreal framework (\url{https://github.com/SurrealAI}) to the research community. We hope that this project could facilitate reproducible research in distributed reinforcement learning.

\subsection*{Acknowledgements}
We would like to thank many members of the Stanford People, AI \& Robots (PAIR) group in using \surreal in their research and providing insightful feedback. This project is supported with computational resources granted by Google Cloud.


\bibliography{surreal}
\bibliographystyle{sysml2019}


\end{document}